\documentclass{article}
\usepackage{ijcai24}

\usepackage{times}
\usepackage{xcolor}
\usepackage{soul}
\usepackage{url}
\usepackage[hidelinks]{hyperref}
\usepackage[utf8]{inputenc}
\usepackage[small]{caption}
\usepackage{natbib}
\usepackage{graphicx}
\usepackage{amsmath}
\usepackage{amsthm}
\usepackage{booktabs}
\usepackage{algorithm}
\usepackage{algorithmic}
\usepackage{todonotes}
\usepackage[switch]{lineno}
\usepackage{enumitem}

\def\name{{\scshape SukhSandesh}}

\linenumbers

\title{\name: An Avatar Therapeutic Question Answering Platform \\for Sexual Education in Rural India}

\author{
Salam Michael Singh$^1$
\and
Shubhmoy Kumar Garg$^3$\and
Amitesh Misra$^{3}$\\
Aaditeshwar Seth$^{3,4,5}$\And
Tanmoy Chakraborty$^{1,2}$
\affiliations
$^1$Department of Electrical Engineering, Indian Institute of Technology Delhi, India\\
$^2$Yardi School of Artificial Intelligence, Indian Institute of Technology Delhi, India\\
$^3$Gram Vaani Community Media (Onion Dev Technologies Pvt. Ltd.), New Delhi, India\\
$^4$Department of Computer Science and Engineering, Indian Institute of Technology Delhi, India\\
$^5$School of Information Technology, Indian Institute of Technology Delhi, India\\
\emails
mike95@iitd.com,
\{shubhmoy.kumar, amitesh.mishra\}@oniondev.com,
aseth@cse.iitd.ac.in,
tanchak@iitd.ac.in
}

\begin{document}
\nolinenumbers
\maketitle

\begin{abstract}
Sexual education aims to foster a healthy lifestyle in terms of emotional, mental and social well-being.  In countries like India, where adolescents form the largest demographic group, they face significant vulnerabilities concerning sexual health. Unfortunately, sexual education is often stigmatized, creating barriers to providing essential counseling and information to this at-risk population. Consequently, issues such as early pregnancy, unsafe abortions, sexually transmitted infections, and sexual violence become prevalent. 
Our current proposal aims to provide a safe and trustworthy platform for sexual education to the vulnerable rural Indian population, thereby fostering the healthy and overall growth of the nation.
In this regard, we strive towards designing \name\footnote{``Sukh'' means pleasure, and ``Sandesh'' means message or communication.}, a multi-staged AI-based Question Answering platform for sexual education tailored to rural India, adhering to safety guardrails and regional language support. By utilizing information retrieval techniques and large language models, \name\ will deliver effective responses to user queries.
We also propose to anonymise the dataset to mitigate safety measures and set AI guardrails against any harmful or unwanted response generation. 
Moreover, an innovative feature of our proposal involves integrating ``avatar therapy'' with \name. This feature will convert AI-generated responses into real-time audio delivered by an animated avatar speaking regional Indian languages. This approach aims to foster empathy and connection, which is particularly beneficial for individuals with limited literacy skills. Partnering with Gram Vaani, an industry leader, we will deploy \name\ to address sexual education needs in rural India.
\end{abstract}

\section{Introduction}
Sexual education strives toward building a person's lifelong sexual health by teaching them attitudes, values, and knowledge about intimacy, relationships, and identity \citep{Ismail2015-vn}. Sexual health is regarded as a state of physical, emotional, mental, and social well-being rather than just the absence of illness or infirmity as reported by WHO\footnote{https://www.who.int/publications/i/item/978924151288}. 
Moreover, sociocultural and psychological factors can increase the likelihood of effectiveness in delivering this education. Sexual education is crucial for adolescents as this is the phase when they undergo hormonal changes\footnote{https://www.who.int/publications/i/item/9789241501552}.
However, sexual education is often stigmatised and thus becomes a hurdle in counselling vulnerable people. Furthermore, this issue is often overlooked by healthcare professionals. For instance, they do not collect a proper history of sexual activities and do not openly discuss the issues due to cultural and traditional norms in society. Notably, there are cases where parents, politicians and teachers have opposed sexual education in Indian states, which include Chhattisgarh, Madhya Pradesh, Rajasthan, Karnataka, Gujarat and Maharashtra \citep{Ismail2015-vn}. They argued that sexual education leads to promiscuity and devalues the Indian ethos and cultural traditions. This will require a tremendous amount of ground-level awareness as such beliefs are deeply rooted in the society irrespective of their background. Moreover, young people are confused by a combination of myths, stigma, ignorance, social disparity, and misinformation. These lead to very critical issues such as sexually transmitted infection (STI), HIV/AIDS, early pregnancy, unsafe abortions, and sexual violence at an alarming rate\footnote{https://www.who.int/publications-detail-redirect/WHO-RHR-10.12}.

\subsection{Sexual Education: An Indian Perspective} Sexual education and sex-related topics are still taboo in India. A study among school students of Ambala District in the state of Haryana in India \citep{Kumar2017-nh} revealed that 93.5\% were in favour of sex education. 86.3\% said sex education could prevent the occurrence of AIDS. 91.5\% of adolescents preferred doctors to give them sex education, followed by teachers. However, many were hesitant to talk with their parents. Another work in the rural region of Bengaluru in India conducted a study among 981 students across 6th to 10th grades \citep{doi:10.1177/09720634231217028}. The study reports that the sexual education program positively impacted the students' confidence. Further, this fostered them towards hygiene and decision-making skills. \citet{doi:10.1016/j.jomh.2012.01.004} conducted a study on the Bengali-speaking Hindu ethnic group from rural and urban demographics. The study was carried out among 101 rural 119 urban adolescent boys. They found that the demographics played a key role in the sexual awareness. The urban boys were more aware, and they were less likely to be involved in unprotected and penetrative sexual activities. Furthermore, the study revealed the role of media and schools in promoting healthy sexual behaviours among adolescents. \citet{GHULE2008167} also found that rural people are more vulnerable towards unprotected sexual activities.

\subsection{Sexual Education is Crucial} For every country, a sexually educated and well-aware population will lead to several benefits such as health, self-confidence and overall development of the economy \citep{david2013sex}. At the global level, adolescents, the most vulnerable group, account for 18\% of the world’s population as of 2009 \citep{Khubchandani2014-gk,Ismail2015-vn}. And India records the largest number of adolescents, with 243 million counts. Because India is a developing country, timely sexual education is crucial for the overall development of the country \citep{Tripathi2013-zu}. If left unattended, this will lead to various risks such as sexually transmitted diseases, premature pregnancies, unsafe abortions, sexual abuse and violence, and overall influence on health and well-being. Hence, there is an urgent need to spread awareness and foster a healthy future for the Indian population \citep{datta2012sex}. Additionally, there are societal stigmas and taboos around sex-related topics in India. Therefore, a ground-level social awareness program is the need of the hour to reduce the stigmas prevailing in sexual education; then, the actual education will be accessible.

\subsection{Existing Studies and Research Gaps} 
This work proposes a multi-staged avatar therapeutic Question Answering (QA) system for sexual education in rural India. Currently, a single system with both avatar therapy and QA for sexual education is not explored. In this section, first, the related work on QA for sexual education is discussed, followed by studies on avatar therapy.

\citet{info:doi/10.2196/29969} developed a chatbot, SnehAI, in Hinglish codemixed data. This platform provides a safe and trustworthy environment for users. The authors targeted adolescent Indian girls to advocate for sexual and reproductive health. \citet{agarwal2021measuring} reported a WhatsApp-based bot named Wulu, which focuses on the gender bias issue in Delhi. A work using a chatbot 
as a sexual health counsellor is reported \citep{info:doi/10.2196/46761}. The chatbot is centred around sexual and reproductive health counselling by facilitating a safe and private platform to build a sense of trustworthiness for vulnerable users. \citet{Nadarzynski2021-lt} conducted a chatbot-based study on women from the UK. They found that most users preferred this chatbot for anonymous education purposes and less suitable where empathy is required. Another chatbot \citep{doi:10.1177/0017896920981122} focused on contraception for young women from Black and Hispanic backgrounds. The study was conducted with 31 women in the age group of 16 to 25 years. The chatbot received a total of 4390 messages. This chatbot shared vital information to boost young women's confidence in contraception. Similarly, \citet{MAEDA20201133} reported a work using a chatbot for fertility awareness. Their awareness program showed significant gains in fertility knowledge.

Avatar-based AI has widely been used in the domains of mental health and cognitive behavioural therapy \citep{info:doi/10.2196/37877, doi:10.1177/0706743719828977, holohan2021like}. \citet{info:doi/10.2196/37877} reported an avatar-based chatbot for mental health anxiety during the COVID-19 pandemic. Their avatar is not a real-time text-to-audio system but an icon to soothe the users. On the other hand, the use of an avatar for combating auditory is used as an avatar therapy for schizophrenia \citep{holohan2021like}. As of now, the avatar has been predominantly used in the mental health domain for therapy purposes, specifically for auditory hallucinations \citep{holohan2021like}.

Hitherto, as discussed above, these chatbots build upon the idea of providing safety and privacy. However, they lack certain aspects, such as these models not being robust to new queries and the responses being fixed as per the training data. Moreover, they lack the scalability for incremental training based on new training instances, and their safety measures are based on user identity anonymisation only without additional guardrails. Furthermore, the avatar is limited to therapy purposes only, while this has several other potential in terms of an empathetic educational bot, which could benefit those who cannot read or write.  

\subsection{Major Contributions} Given the importance of sexual education for a developing country like India, the proposed work plans to develop and deploy \name, an AI-based QA system for sexual education for the rural Indian population. Several risk factors may arise in the absence of well-executed awareness programs and platforms. Also, there are stigmas about sex-related topics, especially in rural India. Thus, keeping this in mind, we aim to address the following objectives:
\begin{enumerate}[noitemsep,nolistsep,topsep=0pt,leftmargin=1em]

    \item \textbf{A Multi-staged AI-based QA System}: We intend to build a QA system which is robust enough to generate responses for existing and new queries. A retrieval module will be used for the existing queries to retrieve the QA pair, which is most similar to the query. Meanwhile, for new queries, generative AI, such as Large Language Models (LLMs), will be used.
    
    \item \textbf{Empathetic Avatar}: We aim to incorporate a 3D avatar that will relay the QA system's response into real-time audio and video with humane facial expressions. This will facilitate empathy and engage the users. Furthermore, the audio-based avatar has the potential to attract rural people out of curiosity. Additionally, a significant portion of the rural population lacks the education level needed for reading and writing. This avatar-based QA system will be effective in this case.
    
    \item \textbf{Regional Language Support}: India is a diverse country with 22 official languages in the Eighth Schedule\footnote{\url{https://www.mha.gov.in/sites/default/files/EighthSchedule_19052017.pdf}}. We intend to support regional languages by using various multilingual and openly available translation APIs.
    
    \item \textbf{Safety and Trustworthiness}: Given the sensitivity and the stigmas surrounding sex-related topics in India, we intend to provide a safe platform that will be non-judgemental, ensure the user's privacy and build overall trust. We aim to approach this by meticulously using safety protocols such as sanitisation and filtering of private information and employing AI guardrails to prevent harmful and unwanted generations.

    \item \textbf{Scalable Model}: We also intend to collaborate closely with Gram Vaani, our industry partner, to gather more new data and incrementally improve and scale the overall performance of \name\ using online learning from the new data over time.
\end{enumerate}

Overall, our proposal addresses multiple dimensions of sustainable development, including health, education, gender equality, reduced inequalities, and partnership building.

\section{Goals}
The primary goal of this work is to provide a safe and secure platform to the vulnerable sections of rural India concerning sexual education where the users can ask, learn and become aware of healthy practices without the fear of getting stigmatised. This work revolves around a multi-staged QA system that aims to build a safe and trustworthy AI system supporting regional language.
\\
\\
\textbf{Multi-staged QA System:} At the global level, a multi-staged QA system for sexual education by leveraging both retrieval and generative AI is less explored. There are reports of the usage of chatbots for sexual health in the UK \citep{Nadarzynski2021-lt}, Bangladesh \citep{10.1145/3411764.3445694}, USA \citep{doi:10.1177/0017896920981122} and South Africa \citep{de2022effects}. However, these systems were solely based on static questionnaires, which lacked empathy and did not consider the contextual information of the overall chat. We intend to exploit the combination of retrieval and LLM-based systems to give responses to even unseen queries based on the contextual information from the database and other information from the user profile for a more personalised and empathetic response. On top of this, we also plan to integrate a 3D real-time avatar to cater for a text-free experience for the users. The avatar will serve two primary purposes: first, to foster empathy through real-time facial and tonal expressions, and second, to provide accessibility to individuals who face challenges with reading or writing, thus catering to the needs of the rural Indian context.  
\\
\\
\textbf{Safety Guardrails}: AI guardrail is a strategy to filter out the model’s input and output. A guardrail takes a set of objects as input. These objects can be the input or output of LLMs. The guardrail then determines if any enforcement actions can be taken. Here, the goal is to reduce the risks embedded in these objects \citep{dong2024building}. Guardrails aim to detect potential misuse during the query stage and work to prevent the model from providing inappropriate answers. AI guardrail is necessary where generative AI is used irrespective of any domain. There is a report \citep{10.1001/jamainternmed.2023.5947} on the use of LLMs for spreading misinformation in the medical domain. The authors used LLMs to generate 102 blogs on vaccines and vaping in just 60 minutes. The blogs targeted multiple diverse societal groups. This raises concern regarding the exploitation of LLMs to spread misinformation on such critical issues. 

Most sexual health-related systems focus on surface-level safety protocols like personal information sanitisation/anonymisation. However, there are other spectrums of safety protocols, such as harmful and unwanted content generation and misinformation. If these protocols are not addressed, it will lead to distrust among the user. Given the sensitivity of sex-related topics in India, it is extremely important to have meticulous and scrutinised guardrails to safeguard the users’ privacy and build trustworthiness.  

\section{Alignment with SDGs}
Our proposal is rightly aligned with the following SDGs:
\begin{itemize}[noitemsep,nolistsep,topsep=0pt,leftmargin=1em]

    \item {\bf Good Health and Well-being (SDG 3):} Sexual education aims to foster a healthy lifestyle, which is crucial for emotional, mental, and social well-being. By addressing issues such as early pregnancy, unsafe abortions, sexually transmitted infections, and sexual violence, sexual education contributes to promoting good health and well-being.

    \item {\bf Quality Education (SDG 4):} We aim to provide a safe and trustworthy platform for sexual education, especially tailored to the rural Indian population. This initiative aligns with the goal of ensuring inclusive and equitable quality education, as it focuses on delivering essential counseling and information to a vulnerable demographic group.
    
    \item {\bf Gender Equality (SDG 5):} Sexual education plays a significant role in promoting gender equality by addressing issues related to sexual and reproductive health, including sexual violence and early pregnancy. By providing access to comprehensive sexual education, especially in regions where adolescents face significant vulnerabilities, this proposal contributes to advancing gender equality.
    
    \item {\bf Reduced Inequalities (SDG 10):} In many countries, including India, access to sexual education is unequal, with rural populations often facing barriers due to stigma and lack of resources. By designing a platform tailored to the needs of rural India and ensuring regional language support, the proposal aims to reduce inequalities in access to essential sexual education resources.
    
    \item {\bf Partnerships for the Goals (SDG 17):} The proposal involves partnering with Gram Vaani for deployment. This collaboration reflects the importance of partnerships between governments, civil society, and the private sector in achieving the Sustainable Development Goals.

\end{itemize}

\section{\name: Model Description}
\begin{figure*}[!t]
\centering
\includegraphics[width=0.9\textwidth]{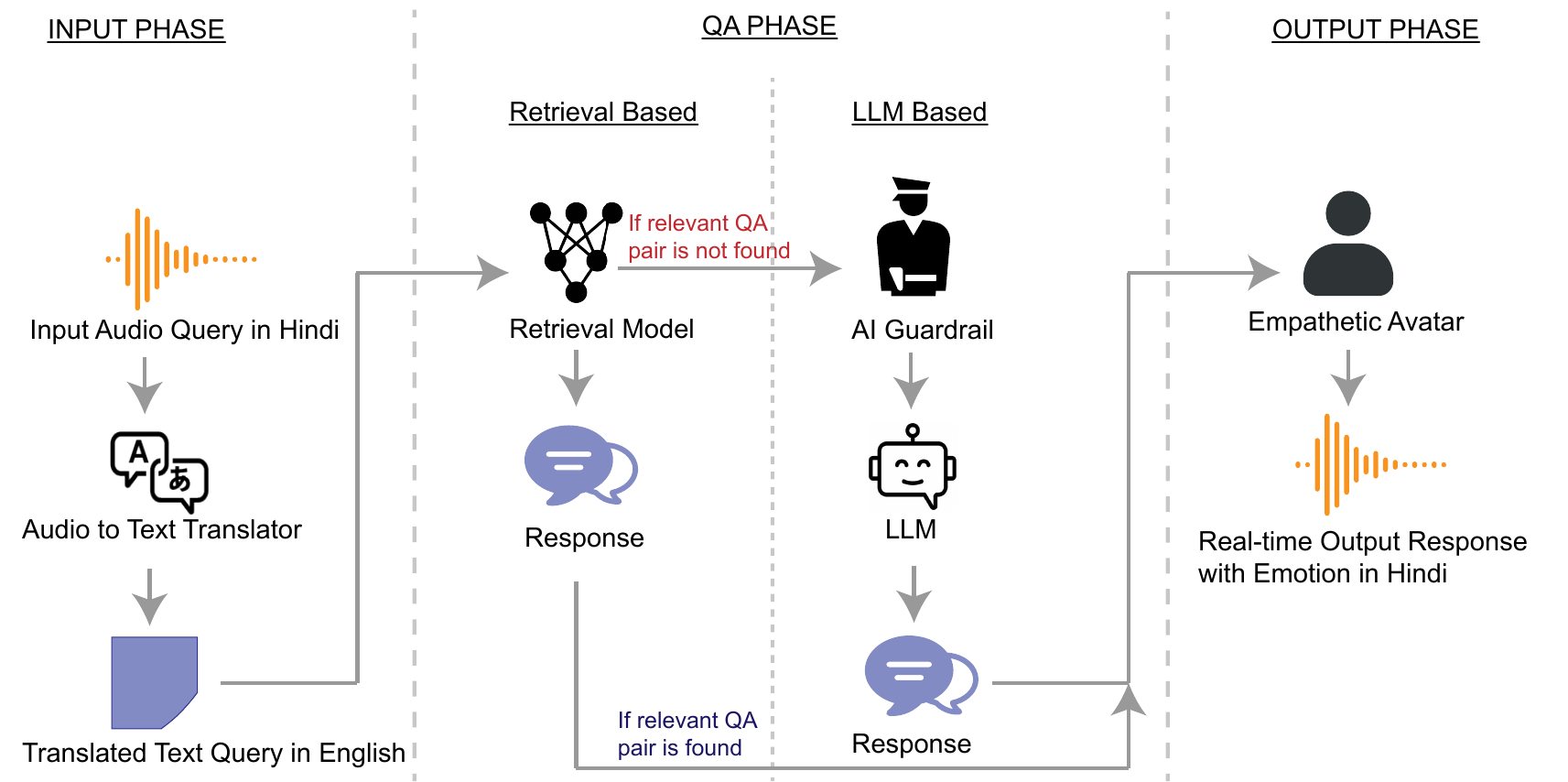}\par 
\caption{A schematic diagram of \name, our proposed multi-staged QA system for sexual education. \name\ consists of three stages, namely, the input phase, the QA phase and the output phase. Initially, it takes the user’s audio query in a regional language, i.e., Hindi, as input. The Hindi audio query is translated to English via the audio-to-text translator API of Bhasini. The text query is then fed into the QA modules. The QA module comprises retrieval-based and LLM-based modules. The retrieval-based module retrieves the most relevant QA pair with the query, and the corresponding answer is returned. If a relevant pair is not found based on the relevance score, LLMs will generate responses with AI guardrails over the query prompt to regularise safety. Finally, the text response is converted to real-time audio-visual representation with emotion using a talking head empathetic avatar. It is to be noted that since the output audio is in Hindi, we aim to finetune the existing talking head avatars using Hindi video data.}
\label{fig:arch}
\end{figure*}

In this section, we elucidate the methodology of \name\ for sexual education in rural India. We propose a multifaceted pipeline towards the QA system by employing various deep learning techniques for retrieval and generation tasks. Furthermore, considering the sensitivity and the taboo around sexual education in rural India, the QA system will incorporate AI guardrails \citep{10.1001/jamainternmed.2023.5947,dong2024building} against sensitive or unwanted behaviours and mock questions. In this section, we will discuss the dataset and individual modules of \name. Figure \ref{fig:arch} presents a schematic diagram of \name.

\subsection{Dataset Sources}
To develop a functional QA system that resonates with the rural Indian demographics, a dataset representing this spectrum is required to make a functional QA system. In this regard, we use a question-answer dataset by Gram Vaani, our industry partner and a non-profit organisation working for rural India on the ground level. The dataset, named KAB (Kahee Ankahee Baatein), contains QA pairs, in which questions sharing the same answer are grouped to represent various ways of asking the same question. The dataset consists of more than 3900 QA entries; it features more than 1200 unique sanitised questions and 3708 unique caller query transcriptions. 

It is categorised into broad themes (such as sexual education, safe sex, sexual intercourse, pregnancy, reproduction, puberty, etc.), enabling more efficient answer retrieval when theme information is available. The dataset is a transcription of a telephonic conversation between the user and the domain expert. The following are some of the features of the dataset:
\begin{itemize}[noitemsep,nolistsep,topsep=0pt,leftmargin=1em]

    \item \textbf{Caller Query}: This field contains the link to the telephonic audio file of the user’s question.
    \item \textbf{Caller Query Transcription}: The text version of the telephonic audio query after the transcription.
    \item \textbf{Relevant Question}: Callers often include personal and irrelevant information in their questions. This irrelevant information is removed from these questions to create the relevant question.
    \item \textbf{Sanitised Question}: The question contains some sensitive information sanitised to create a refined version. These sanitised questions can be mapped to the corresponding relevant questions.
    \item \textbf{Theme and Sub-Theme}: The caller tags all the query transcriptions with a corresponding theme and sub-theme.

{We propose to leverage the sexual education theme from this dataset for the QA system. The answers will be retrieved for the existing related questions, while new questions will be fed to the LLMs to generate the answers. } Meanwhile, we will augment KAB with more human-generated questions and answers received from health workers, which will be incorporated into the model over time.

\end{itemize}

\subsection{Audio-to-Text Conversion of Telephonic Transcripts}
The input to \name\ will be a user-generated audio query. However, the intermediate QA generation from both the retrieval and generative AI-based approach requires in-text format as these are text-only models. The source of the KAB dataset is the audio recordings from the Interactive Voice Response (IVR) of the users in Hindi. These audio recordings in Hindi are directly translated into English text via Bhasini's audio-to-text translation API \footnote{\url{https://anuvaad.bhashini.gov.in/}}. For training and inferring the intermediate QA modules, only the text is required. During the inference, initially, the user will give input via audio. This makes the platform accessible to even the underprivileged section of society. This audio query in Hindi will then be translated into English text via the Bhasini API, and the QA modules will generate the response. Finally, the talking head avatar animates the text response into real-time video with emotion. However, most avatar models are in English; therefore, we need to finetune them with Hindi video data. Additionally, the input to \name\ is Hindi audio, and the final desirable output from the empathetic avatar is Hindi audio. Therefore, instead of translating the Hindi audio into English text and then feeding the text into an English-only LLM, we plan to simply recognise the Hindi text using Bhasini's automatic speech recognition (ASR) API. This will reduce translation-related errors and maintain the demographic and cultural tones. Furthermore, this Hindi text can be directly fed to Hindi LLMs \citep{gala2024airavata} or multilingual LLMs \citep{workshop2022bloom}. 

\subsection{Retrieve Previous Questions}
The KAB dataset, compiled by grassroots NGOs and self-help groups (SHGs), reflects the demographic makeup of rural India. As a result, we anticipate encountering numerous new queries that resemble existing ones. To address this, we will leverage these related queries through a retrieval-based system. This approach involves retrieving answers for the most pertinent queries based on their relevance to existing ones.
Initially, we will explore both the dense \citep{10.1162/tacl_a_00369, izacard2021unsupervised} and sparse retrieval \citep{10.1145/1031171.1031181} approaches and select the optimal one. As we need to retrieve only the relevant QA pair, it is crucial to quantify the relevance of the retrieved document. Furthermore, {we will explore the re-ranking-based approach \citep{sung-etal-2023-optimizing, fajcik-etal-2021-r2-d2} for quantifying the relevance score and come up with some threshold value to consider or discard the retrieved QA pair.}

\subsection{QA Generation from LLMs}
Retrieval-based QA is effective when a similar query is already present and does not account for the client's nuances and preferences. However, LLMs have shown their efficacy in flexible QA generation tasks \citep{10.1162/tacl_a_00407} by considering the contextual information. The LLM generation is typically performed in the following ways. (i) \textbf{Finetuning}: The existing QA dataset can be used to finetune the LLM using a supervised finetuning \citep{NEURIPS2022_b1efde53} approach. For a decoder-only LLM such as Llama \citep{touvron2023llama}, Bloom \citep{workshop2022bloom}, Mistral \citep{jiang2023mistral}, 
OpenHathi\footnote{https://www.sarvam.ai/blog/announcing-openhathi-series}, and Airavata \cite{gala2024airavata}, the question and answers are concatenated into a single string, and the model is tasked to generate the next token in an autoregressive manner. Bloom is a multilingual LLm among these LLMs, while OpenHathi and Airavata are LLMs in Hindi. Meanwhile, for an encoder-decoder LLM such as T5 \citep{10.5555/3455716.3455856}, mT0 \citep{muennighoff-etal-2023-crosslingual} and Aya \citep{ustun2024aya}, the question is treated as the source and the answer as the target. Here, T5 is an English-only LLM, while mT0 and Aya are multilingual models supporting Hindi. (ii) \textbf{In-context Learning:} This approach does not train or finetune the existing LLM \citep{dong2022survey}. Rather, it exploits the generalisability of the model \citep{min-etal-2022-rethinking} via prompting with examples of the question-answer pairs to guide the generation of the LLM.  
{We propose to utilise the generative prowess of LLMs (viz., encoder-decoder and decoder only) for QA generation for sexual education in rural India. In doing so, we will also explore fine-tuning and in-context learning approaches.}

\subsection{Empathetic Avatar}
Avatar-based chatbots and systems have been predominantly used as therapeutic modules \citep{info:doi/10.2196/37877, doi:10.1177/0706743719828977, holohan2021like,info:doi/10.2196/37877}. In this work, our goal is to use a 3D avatar. \textit{Specifically, we propose to use AI talking head avatar \citep{weisman2021face, zhen2023human} which enhances the human-computer interaction.} These AI talking head avatar converts text into speech and simultaneously animate the avatar by synthesising lip motion sequences corresponding to the text \citep{zhen2023human}. The overall output is a real-time video where the avatar seems to talk like a human with visual and tonal emotions according to the context. 

In our pretext, the avatar will convert the text responses generated by the QA system in real time. These responses can be from either the retrieval-based method or LLMs. The conversion will result in an audio-visual interpretation of the text response. The avatar will try to build empathy by showing real-time, humane facial expressions according to the context.  
We also explore the opportunity to extend the system further by providing audio-based queries instead of text so that the users won't need to type any text. Moreover, a significant portion of the Indian rural population lacks basic education, hindering them from reading or writing abilities. This text-free avatar-based QA system will be of great potential as it won’t require the users to have reading or writing abilities, and their curiosity to use the avatar will increase the overall usage of the QA system for sexual education.

\subsection{Support to Regional Languages}
One of the major challenges for a QA system in the rural Indian setting is the language barrier. India is a diverse country with multiple regional languages and dialects. A large section of the rural population is not fluent or does not know English. However, most existing LLMs are English-centric and will be ineffective for regional languages. {To address this, we aim to generate the Hindi QA dataset}. However, this would require a machine translation (MT) system trained using the regional language. In this regard, we will only focus on Hindi as the choice of the regional language as it is spoken and understood by a majority of the Indian population, and there are already existing powerful MT systems for English-Hindi pairs \citep{kunchukuttan-etal-2018-iit} and abundant parallel corpus. We also plan to extend this to other popular Indian languages, such as Tamil and Bengali. However, existing MT systems cannot generate sensitive and technical terms like reproductive organs in Hindi. We will also explore the domain-specific perspective of the model. Additionally, instead of utilising an English-centric LLM and postprocessing the QA generation into the regional language, we propose extending the LLM into the regional language for our specific task.

\subsection{Guardrails Against Sensitive or Unwanted Behaviours} 
Current LLMs are extremely powerful, but can be adjusted with prompts. This flexibility of LLMs comes with several risks, but not limited to sensitive data leaks, abusive responses, immoral queries, and misinformation that malign the integrity of the AI system and have a serious impact on users and societies \citep{dong2024building,10.1001/jamainternmed.2023.5947}. In the context of our proposal, sexual education remains a taboo subject, leading individuals to shy away from openly expressing their views or engaging in discussions about it. Hence, it is critical to sanitise or anonymise the identity and filter out any personal information in the dataset to increase the trustworthiness and a sense of privacy among the users. Additionally, the LLM-based QA system may generate undesired behaviours such as misinformation and toxic, insensitive, vulgar and harmful content. \textit{We aim to construct guardrails against these potential risks to promote a safe and credible AI system for QA generation.} Particularly, we will finetune the model with some instruction prompts where we will explicitly instruct to generate harmless responses. The initial guardrail will be a meticulously chosen prompt and later on, we will explore additional advanced guardrails.

\section{Deployment Plan with Gram Vaani}

{\bf Existing Capabilities of Gram Vaani.} Gram Vaani currently operates the Kahi Ankahi Baatein (KAB) service over an IVR system. A toll-free phone number was publicized via a network of partners working in the areas of gender equality, sexual and reproductive health, grassroots media, etc. Users can give a call to this phone number and cut the call. The IVR server then automatically calls them back within a few seconds. Over this voice call, users can use keypresses to navigate across published audio messages, ask questions by recording a voice message, and listen to answers to questions asked by others. Incoming questions are heard by a team of trained content moderators. The KAB service started in 2015 and receives an average of 300 calls per day from adolescents and young couples. New questions or topical questions are put up to a gynaecologist partner who has been associated with the KAB service for several years; she provides voice-recorded answers to the questions, which are published on the KAB IVR in the following week. 

{\bf \name's Deployment Strategy.} As presented in this proposal, the KAB service is being enhanced to also operate over a smartphone chatbot and to use the new capabilities of generative AI to improve the interactivity of the service. The service will also be linked via the existing KAB IVR and publicized through two further venues. First, Gram Vaani runs a community volunteer network in more than 1000 panchayats in Northern India (states of Bihar, Jharkhand, Uttar Pradesh, and Madhya Pradesh); the volunteers assist communities with getting access to social entitlements under government schemes, get answers to queries by farmers related to agriculture, learn about digital financial services and online safety, etc. \citep{moitra2016design}. Over 40\% of the volunteers are women. This face-to-face connection by the volunteers will be used to publicize the service further through local youth networks. second, in these panchayats, Gram Vaani operates an IVR service similar to KAB but with more generic content related to hyperlocal news, health, and social issues such as early marriage and domestic violence etc. Over 100,000 unique users access the service each month, about 30\% of whom are youth under 25 years of age \citep{seth2020reflections}. We will also publicize the chatbot via this service. Overall, we expect to achieve a usage of more than 1000 users per day of the chatbot service. With regular updates, maintenance, and user feedback, we will ensure that the system remains secure, stable, user-friendly, and continually meets the needs of its users.

\section{Model Evaluation}
We plan to evaluate the performance of \name\ based on multiple factors as mentioned below:
\begin{itemize}[noitemsep,nolistsep,topsep=0pt,leftmargin=1em]

    \item \textbf{Precision and Accuracy}: The model should be able to generate an adequate and fluent answer that the gold labels will gauge. Evaluation metrics such as BLEU \citep{papineni-etal-2002-bleu}, ROUGE \citep{lin-2004-rouge}, BERTScore \citep{zhang2019bertscore} can be utilised.
    
    \item \textbf{Bias}: The model should be agnostic towards any bias and should be fair and impartial in its generation.
    
    \item \textbf{Extent of Hallucination}: As a generative AI model, the proposed system may be prone to hallucination. However, the model should only generate adequate responses and reduce hallucinations. For this aspect, Chainpoll \citep{friel2023chainpoll} can be used as an evaluation metric.
    
    \item \textbf{Generalisability}: The model should be able to generalise well to new data that it has not seen before. For this, we perform any precision and accuracy evaluation metric on the new data. 
    
    \item \textbf{Interpretability}: The model should be interpretable for generation. This will facilitate causal relations between the input features of the model and the overall QA generations.
    
    \item \textbf{Robustness}: The model should be robust to changes in the data and to any noise or errors that may be present. The proposed method is a pipeline of multiple units; any errors propagated in the channel should not affect the overall performance.
    
    \item \textbf{Scalability}: The model should be scalable to efficiently handle large volumes of data. And it should be able to learn with new data incrementally.

\end{itemize}
\section{Challenges}
An AI-based QA system like \name\ poses multiple challenges. These challenges get multiplied when applied to a rural Indian pretext. The following are some of the challenges of a QA system for sexual education in rural India:
\begin{itemize}[noitemsep,nolistsep,topsep=0pt,leftmargin=1em]

    \item \textbf{Taboos and Stigmas Related to Sex}: A large number of the rural population in India is stigmatised from openly speaking and sharing their view on sex-related topics, specifically females. This hinders the data acquisition step, where the users are expected to share and ask educational questions related to sex. Furthermore, even if the model is deployed, then also the stigma will prevail with using a sexual education tool. Moreover, many people normalise or do not consider the problems as problems such as having more than four children or not sending the girl child to school. Therefore, a substantial effort at the ground level for awareness is required, which we expect to overcome with Gram Vaani's ground-level workers
    \item \textbf{Lack of Labeled Data}: LLMs necessitate a significant volume of data for training to achieve satisfactory performance. However, acquiring sex education-related data, particularly within the rural Indian context, is challenging due to the taboo surrounding topics related to sexuality. Consequently, gathering sufficient data poses a major challenge. Moreover, \name\ will require machine translation or alternative multilingual strategies to accommodate regional languages. Nevertheless, the availability of such multilingual resources is also constrained.
    \item \textbf{Data Quality}: Sex education-related data can be subject to quality issues such as gender bias, normalising taboos, or misaligned instances in the parallel corpus for the machine translation phase, leading to biased results, or the data may not be representative of the population being studied. And, as our pipeline requires speech-to-text conversion, there might be cases where the error in the conversion might get propagated further downstream. 
    \item \textbf{High Carbon Footprint}: LLMs require a significant amount of energy, which contributes to carbon footprints. This impact can adversely affect Sustainable Development Goal 7 (Affordable and Clean Energy) and Sustainable Development Goal 13 (Climate Action).
    \item \textbf{Lack of Skilled Personnel}: AI model training and deployment require skilled manpower. Additionally, these personnel must be adaptive to the growing pace of AI. Nonetheless, there is a need for more skilled personnel in AI in India.
\end{itemize}

\section{Risks, Limitations, Ethical Considerations}
The taboos and stigmas w.r.t. sex-related topics prevail among the rural Indian population. Until and unless ground-level awareness is carried out, it will be a limitation and hurdle in the overall project. Furthermore, the use of LLMs comes with the issue of hallucination. We aim to address this by forcing the LLMs to generate only factually correct responses; however, the model may not be fully hallucination-proof. We aim to use a 3D avatar for real-time text-to-speech interpretation with emotions. However, perfect emotions with facial structure will be difficult to mimic. Furthermore, if done wrongly, it will have an uncanny behaviour which might be uncomfortable sometimes. 
We also plan to incorporate AI safety guardrails into the model to protect and filter out any unwanted and harmful content. However, it is important to note that these guardrails may be tweaked because designing robust guardrails against any new adversaries is challenging. Finally, there might be some ethical considerations in data acquisition and their use, particularly sensitive user information. This will require their consent and sanitisation of the data to filter out and anonymise their information.

\section{Expected Results and Long-Term Plans}
It is expected that the comprehensive findings of the planned study on sexual education in rural India will offer a deep understanding of the complex issues surrounding sex. The project intends to address societal biases towards sex by identifying the factors that contribute to the stigmas associated with sexual education. It is anticipated that the adoption of a customised, private quality assurance system will increase target users' trustworthiness and confidence. Long-term plans also call for creating a personalised education ecosystem. This will meet the changing needs of the Indian rural population. Furthermore, community-based engagement and awareness programmes will make a secure platform for sexual education available. Lastly, by utilizing target profiles, the integrated results can offer different perspectives on spreading awareness.

\section{Conclusion}
Sexual-related issues constitute a significant concern in rural India. People are often unaware of sexual education due to the stigmas and taboos prevailing in sex. The subject is not openly discussed, and those who attempt to address it are often met with social disapproval. This raises a pressing concern, which is evident with the normalisation of child marriage, unprotected sex, unsafe abortions, lack of menstrual sanitation and maternity care. To address these issues, our proposal aims to provide a safe and private space for this vulnerable and marginalised section of the Indian population. Specifically, we propose building \name, a QA system for sexual education in the rural Indian setting by leveraging the prowess of traditional AI as well as generative models.  
Moreover, for enhanced accessibility and user convenience, we also plan to use a real-time avatar-based QA system. This system will prioritise empathy, where the input query will be audio-based in Indian languages. The responses generated will be the real-time audio-visual representation of the text response from the QA system. The system will be scalable and continuously adaptive. \name\  will incorporate new training data and get updated over time. Additionally, to expand the outreach, \name\ will also support regional languages. Finally, we will also integrate AI guardrails for security and privacy concerns to prevent harmful and unwanted responses. In collaboration with Gram Vaani, we will deploy \name\ in rural India as a mobile-based application to spread widespread awareness of sexual education, which is still an abandoned and controversial topic in India.

\clearpage





{\small 
\bibliographystyle{named}
\bibliography{ijcai24}}

\end{document}